\documentclass[final]{cvpr}

\usepackage{times}
\usepackage{epsfig}
\usepackage{graphicx}
\usepackage{amsmath}
\usepackage{amssymb}
\usepackage{bbm}
\usepackage{multirow}
\usepackage{multicol}
\usepackage{amsmath}
\usepackage{comment}
\usepackage{float} 
\usepackage{amssymb}
\usepackage{pifont}

\usepackage{colortbl}
\def\greyColor{\textcolor{pblue}}
\usepackage{color}
\definecolor{pblue}{RGB}{0,0,0}
\definecolor{grey}{RGB}{232,232,232}

\newcommand{\tablestyle}[2]{\setlength{\tabcolsep}{#1}\renewcommand{\arraystretch}{#2}\centering\small}
\newcommand{\tabincell}[2]{\begin{tabular}{@{}#1@{}}#2\end{tabular}}  
\newlength\savewidth\newcommand\shline{\noalign{\global\savewidth\arrayrulewidth
  \global\arrayrulewidth 1pt}\hline\noalign{\global\arrayrulewidth\savewidth}}
\definecolor{forestgreen}{RGB}{47, 159, 87}
\newcommand{\cmark}{\color{forestgreen}\ding{51}}%
\newcommand{\xmark}{\color{red}\ding{55}}%

\usepackage[pagebackref=true,breaklinks=true,colorlinks,bookmarks=false]{hyperref}

\begin{document}

\title{ParamCrop: Parametric Cubic Cropping for Video Contrastive Learning
\vspace{-0.4cm}}

\author{Zhiwu Qing$^{1,3\dag}$ \quad Ziyuan Huang$^{2,3\dag}$ \quad Shiwei Zhang$^{3*}$ \quad Mingqian Tang$^3$ \\
\quad  Changxin Gao$^1$ \quad Marcelo H. Ang Jr$^2$   \quad Rong Jin$^3$   \quad Nong Sang$^{1*}$
\\
$^1$Key Laboratory of Image Processing and Intelligent Control \\ School of Artificial Intelligence and Automation, Huazhong University of Science and Technology\\
$^2$ARC, National University of Singapore\\
$^3$Alibaba Group\\
{\tt\small \{qzw, cgao, nsang\}@hust.edu.cn}\\
{\tt\small ziyuan.huang@u.nus.edu, mpeangh@nus.edu.sg}\\
{\tt\small \{zhangjin.zsw, mingqian.tmq, jinrong.jr\}@alibaba-inc.com}
\vspace{-0.4cm}
}

\maketitle

\let\thefootnote\relax\footnotetext{$\dag$ Equal Contribution.}
\let\thefootnote\relax\footnotetext{$*$ Corresponding authors.}
\let\thefootnote\relax\footnotetext{This work is done when Z. Qing, Z. Huang are interns at Alibaba Group.}

\begin{abstract}
\vspace{-3mm}
The central idea of contrastive learning is to discriminate between different instances and force different views from the same instance to share the same representation. To avoid trivial solutions, augmentation plays an important role in generating different views, among which random cropping is shown to be effective for the model to learn a generalized and robust representation. 
Commonly used random crop operation keeps the distribution of the difference between two views unchanged along the training process.
In this work, we show that adaptively controlling the disparity between two augmented views along the training process enhances the quality of the learned representations.
Specifically, we present a parametric cubic cropping operation, ParamCrop, for video contrastive learning, which automatically crops a 3D cubic by differentiable 3D affine transformations.
ParamCrop is trained simultaneously with the video backbone using an adversarial objective and learns an optimal cropping strategy from the data.
The visualizations show that ParamCrop adaptively controls the center distance and the IoU between two augmented views, and the learned change in the disparity along the training process is beneficial to learning a strong representation. 
Extensive ablation studies demonstrate the effectiveness of the proposed ParamCrop on multiple contrastive learning frameworks and video backbones. 
Codes and models will be available.
\end{abstract}

\begin{figure}[t]
\begin{center}
   \includegraphics[width=1.0\linewidth]{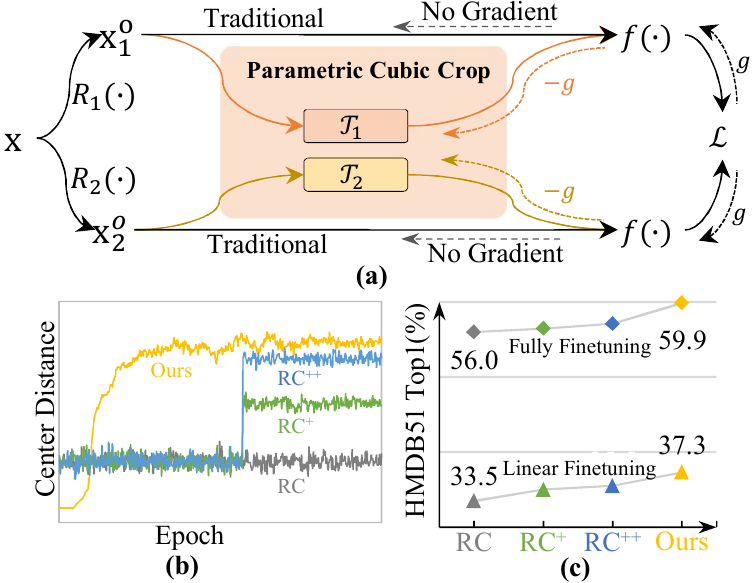}
\end{center}
\vspace{-4mm}
   \caption{(a) Comparison between our proposed ParamCrop framework and the traditional random cropping in the constrastive framework. `$g$' is normal gradient, and `$-g$' indicates reversed gradient.
   The parametric cubic crop in ParamCrop is connected after other random transformations $R_1, R_2$ to adaptively crop two augmented views during the training process.
   (b) (c) Preliminary experiments show that increasing the mean center distance (\ie, RC$^+$ and RC$^{++}$) between two cropping regions in the later stage of contrastive training can benefit the learned representations.
   }
\label{fig:first_fig}
\vspace{-3mm}
\end{figure}

\section{Introduction}
Learning representations from massive unlabeled data is a prominent research topic in computer vision for reducing the need for laborious and time-consuming manual annotations~\cite{chen2020simclr,he2020moco,swav,byol,simsiam}. 
In the video analysis paradigm, which is the focus of our work, such unsupervised learning strategies are more crucial because of their increased labeling difficulty caused by the ambiguous association between videos and their labels.
Early works manually design proxy tasks for learning videos, either by generalizing methods from the image domain~\cite{jing2018-3drotnet,kim2019cubic_puzzles} or by exploiting temporal properties of videos~\cite{xu2019vcop,luo2020vcp,diba2019dynamonet,benaim2020speednet,misra2016shuffle_t_order,zhukov2020learning_t_order,lee2017unsupervised_t_order}, where visual structures and contents are learned through solving these proxy tasks. 
Inspired by the instance discimination task~\cite{wu2018ins_dis}, contrastive based self-supervised approaches have achieved impressive performances~\cite{han2019dpc,han2020memdpc,versatile,hu2021corp,feichtenhofer2021largescale,qian2021cvrl}, which shows its potential to learn advanced semantic information from unlabelled videos.
One of the key factors in the success of the contrastive learning framework is data augmentation~\cite{qian2021cvrl,chen2020simclr} for different views of the same instance, which prevents the model from trivial solutions.
Among common data augmentation strategies, it is shown in ~\cite{chen2020simclr} that random cropping is one of the most effective operations. 

Most current approaches for contrastive learning use random cropping with completely random spatio-temporal location selections. 
It keeps the distribution of the difference between two cropped views unchanged along the training process, 
as showcased in grey (\ie, RC) in Figure~\ref{fig:first_fig}($b$). 
Inspired by curriculum learning~\cite{bengio2009curriculum_learning}, we slightly increase the difficulty in the later stage of contrastive training by increasing the central distance between two cropped views (Figure~\ref{fig:first_fig}(b)). 
Figure~\ref{fig:first_fig}(c) shows that this yields a stronger representation for the downstream action recognition task.

Motivated by this, we propose to \textit{adaptively control the disparity between two views} for contrastive learning.
Specifically, we present a parametric cubic cropping dubbed ParamCrop, where cubic cropping refers to cropping a 3D cube from the input video. The central component of ParamCrop is a differentiable spatio-temporal cropping operation. This enables ParamCrop to be trained simultaneously with the video backbone and adjust the cropping strategy on the fly. 
The objective of ParamCrop is adversarial to the video backbone, \ie, to increase the contrastive loss. 
Hence, initialized with the simplest setting where two cropped views largely overlap, ParamCrop gradually increases the disparity between two views. 
Further, we introduce an early stopping strategy for ParamCrop to control the maximum disparity, since there exists a sweet spot in the intensity of augmentations for contrastive learning~\cite{tian2020whatmakes}.
Compared to the auto augmentation approaches in the supervised setting~\cite{lim2019fastaugment,cubuk2019autoaugment,ho2019pba,cubuk2020randaugment}, our objective is radically different. The auto augmentation methods aim to increase the data diversity, while ParamCrop sets out to discover an optimal cropping strategy to reasonably control the differences between two views along the training process.

We quantitatively evaluate the representations trained by ParamCrop on two downstream tasks, \textit{i.e.,} video action recognition and video retrieval.
Notable improvements are observed on multiple mainstream contrastive learning frameworks and video backbones, which shows that the idea of adaptively increasing the disparity between two views along the training process is crucial to learning generalized representations.

\textbf{Contributions.} 
\textbf{(a)} 
We propose a cropping strategy that adaptively controls the disparity between two cropped views along training process;
\textbf{(b)} 
We propose a differentiable cropping method that can be trained end-to-end together with the backbone;
\textbf{(c)} Extensive experiments on multiple downstream tasks and datasets demonstrate the effectiveness our cropping strategy.

\begin{figure*}[t]
\begin{center}
   \includegraphics[width=1.0\linewidth]{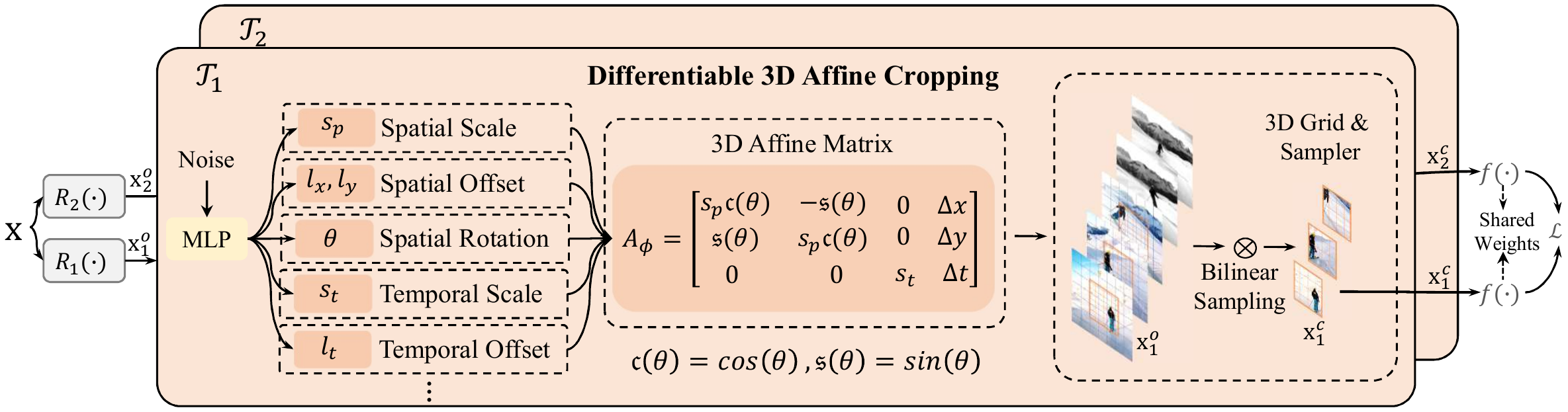}
\end{center}
\vspace{-4mm}
   \caption{
   The overall framework of the proposed ParamCrop and detailed illustration of the differentiable spatio-temporal cropping operation workflow. 
   Two independent cropping modules $\mathcal{T}_1$ and $ \mathcal{T}_2$ are inserted between other random augmentations $\{R_1, R_2\}$ (if any) and the backbone$f$.
   }
   \vspace{-4mm}
\label{fig:fw}
\end{figure*}
\section{Related Work}
\textbf{
Self-supervised video representation learning.
}
To avoid the laborious and time-consuming annotation process, a wide range of prior works have proposed different approaches for leveraging unlabelled data. 
Recent endeavors can be mainly divided into two categories, that is pretext task based approaches~\cite{jing2018-3drotnet,benaim2020speednet,diba2019dynamonet,wang2020video_peace,luo2020vcp,xu2019vcop} and contrastive learning based ones~\cite{han2019dpc,han2020memdpc,coclr,chen2020simclr,he2020moco,oord2018cpc}. The former ones usually introduce a proxy task for the model to solve. Besides the simple generalization from the image domain~\cite{rotation,noroozi2016jigsaw} such as rotation prediction~\cite{jing2018-3drotnet} and solving puzzles\cite{kim2019cubic_puzzles,luo2020vcp}, other tasks include predictions on the temporal dimension such as speed prediction~\cite{benaim2020speednet}, frame/clip order prediction~\cite{lee2017sort_seq,xu2019vcop}, and predicting future frames~\cite{diba2019dynamonet}, \textit{etc.} Closely related to our work is contrastive learning based approaches, which were inspired by the instance discrimination task~\cite{wu2018ins_dis}. It requires the model to discriminate augmented samples from the same instance from other instances and map different views of the same instance to the same representation. Based on the formulation in \cite{oord2018cpc}, \cite{han2019dpc,han2020memdpc} contrast between the representation of the predicted future frames and that of the real ones. Some recent works exploit video pace variation as augmentation and contrast between representations with different paces~\cite{wang2020video_peace,chen2020rspnet}. Whether it is in the video paradigm, which is the focus of this paper, or in the image domain, augmentations are all shown to be critical to learning a strong representation. 
Yet all of them apply random cropping with completely random spatio-temporal location and uniform scale variation parameters along the whole training process. 
We build our approach for video contrastive learning upon the simplest contrastive framework~\cite{chen2020simclr,he2020moco} and show that parameterized cubic cropping controlling the change process is conducive to the improvement of learned representations.

\textbf{Data Augmentation.}
The importance of the data augmentation has already been discovered in the supervised learning. The main objective of data augmentation in supervised settings is  to enhance the diversity of the training data so that the model can learn generalized representations. The hand-craft data augmentations confuse the network by erasing information~\cite{random_erasing, cutout} or mixing difference samples~\cite{zhang2017mixup,hendrycks2019augmix}. To reduce the dependence on human expertise,  automatic data augmentations are proposed to search the combination of augmentation policies by undifferentiable methods ~\cite{cubuk2019autoaugment,ho2019pba,lim2019fastaugment} or online learnable strategy~\cite{li2020dada}. Although our approach is similar to automatic augmentation, our ParamCrop learn the cropping region that adaptive to the training process, rather than the combination of augmentations.
Further, in unsupervised learning, Alex \textit{et al.}~\cite{tamkin2020viewmaker} propose a learnable color transformation improve the robustness of the network. However, this work explore the automatic cropping operation to provide adaptive augmented views for video contrastive learning.

\section{Method}

This section introduces the proposed parametric cubic cropping framework, ParamCrop, based on contrastive learning.
The objective of ParamCrop is to adaptively control the cropping disparity between two generated views during the contrastive training process. 
To this end, we propose a differentiable spatio-temporal cropping operation, which can utilize the cropping parameters regressed by the cropping networks(\ie, an MLP for regressing the cropping parameters) to realize the 3D cropping operation.
This enables the cropping operation to be jointly optimized with the video backbone. Two identical but independent cropping modules are connected respectively to each of the views. To gradually increase the disparity between views in training, we first initialize the cropping networks so that two crops share a similar space-time location, and then train the networks adversarially along with the video backbone using a gradient reversal strategy. The overall framework, as well as the detailed workflow, is visualized in Figure~\ref{fig:fw}. 

\subsection{Contrastive Learning}

Our ParamCrop framework is built upon recent simplified contrastive learning frameworks~\cite{chen2020simclr,he2020moco}, where the model is trained to maximize the agreement between two augmented views of the same instance and minimize that from different instances. 
Suppose there are $N$ different samples, we can generate $2N$ augmented views and the contrastive loss can be written as:
\begin{equation}
    \mathcal{L}=\frac{1}{2N}\sum_{k=1}^{N}[\ell(2k-1, 2k) + \ell(2k, 2k-1)] \ ,
\end{equation}
where $\ell(i, j)$ defines the loss between two paired samples: 
\begin{equation}
    \ell(i, j)=-\text{log}\frac{\text{exp}(\mathbf{c}_{i,j} / \tau)}{\sum_{k=1}^{2N}\mathbbm{1}_{[k\ne i]}\text{exp}(\mathbf{c}_{i,k}/\tau)}\ ,
\end{equation}
where $\mathbf{c}_{i,j}$ is the cosine similarity between the representation of view $i, j$, and $\tau$ is the temperature parameter.
Typically, the two views are generated by the same set of augmentation, usually consisting of random cropping, color jittering, \textit{etc.}
In standard contrastive methods, the augmentation strategy keeps unchanged along the training process. 
Hence, the distribution of the view difference is consistent along the training process.

\subsection{Differentiable 3D Affine Cropping}

For the proposed ParamCrop to control the cropping disparity during the training process, the cropping operation is firstly required to be trainable. 
Inspired by STN~\cite{stn}, we extend the image affine transformations to video 3D affine transformations for cropping cubes from original videos in a differentiable way.

\begin{figure}[t]
\begin{center}
   \includegraphics[width=0.9\linewidth]{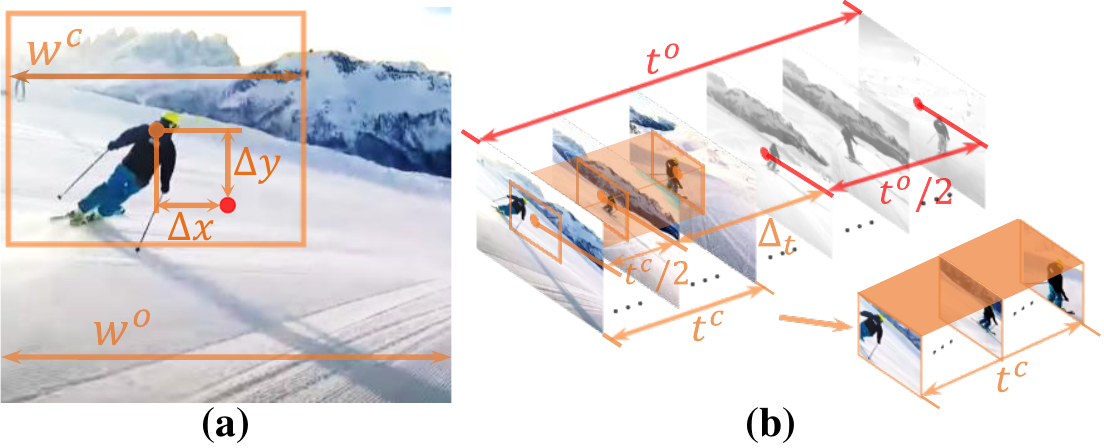}
\end{center}
\vspace{-4mm}
   \caption{The diagram of preliminary parameters for the 3D affine transformation. 
   \textbf{(a)} 
   Illustration of the cropped region with parameters $w_c, w_0, \Delta x, \Delta y$.
   %
   %
   \textbf{(b)} Illustration of the temporal length of original video: $t_o$, cropped video: $t_c$.
   %
   }
\label{fig:st}
\vspace{-4mm}
\end{figure}

\noindent\textbf{Cubic cropping with 3D affine transformations. }Before introducing the 3D affine transformation, we first define mathematical notations $\mathbf{x}^o$, $\mathbf{x}^c$ as the original videos and the cropped videos, respectively. Then we further define the original video width $w_o$, the cropped video width $w_c$, the temporal length of the original video $t_o$ and the cropped video $t_c$. Figure~\ref{fig:st} illustrates their meanings intuitively.

With these notations, a 3D affine transformation matrix $A_\phi$ for calculating the transformation relationship from the homogeneous coordinate in the cropped video to the original video can be defined as follows: 
\begin{equation}
A_{\phi}=
    \left[\begin{array}{cccc}
    s_p\textit{cos}(\theta) & -\textit{sin}(\theta) & 0 & \Delta x \\
    \textit{sin}(\theta) & s_p\textit{cos}(\theta) & 0 & \Delta y \\
    0 & 0 & s_t & \Delta t
    \end{array}\right]\ ,
\end{equation}

\noindent where $s_p=w_c/w_o$ is the region scale, $\theta$ refers to the spatial rotation angle, $(\Delta x, \Delta y)$ indicates the spatial center position offsets, $s_t=t_c/t_o$ is the temporal scale and $\Delta t$ means the temporal offset. 
Since we only implement cropping operations, irrelevant parameters in $A_\phi$ are set to 0 by default.
Therefore, there are altogether six parameters that can be learned in the affine matrix for 3D cropping operation.

\begin{figure}[t]
\begin{center}
   \includegraphics[width=0.9\linewidth]{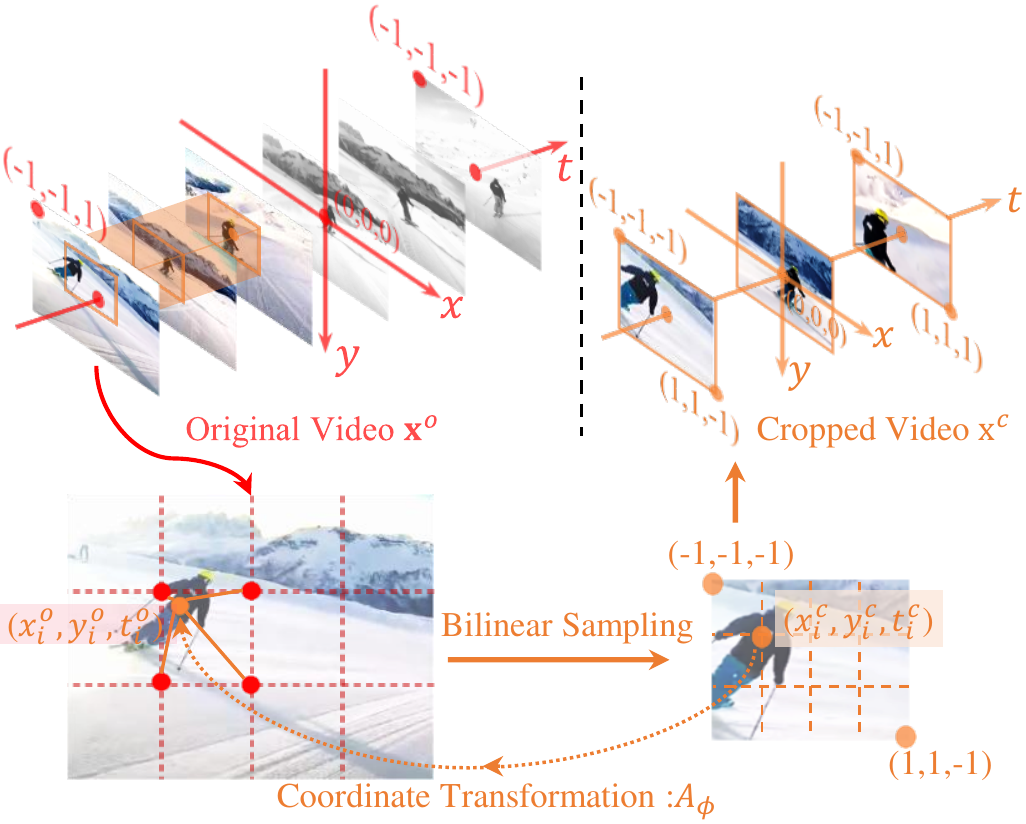}
   \vspace{-5mm}
\end{center}
   \caption{
   Illustration of coordinate systems and the cropping process.
   In the coordinate systems, each pixel location in the $x,y,t$-axis is normalized to within the range of [-1, 1].
   For the cropping process, the uniform distributed coordinates in the cropped video $\mathbf{x}^c$ are first transformed to the original video $\mathbf{x}^o$ using 3D affine transformation matrix $A_\phi$.
   Then a bilinear interpolation is performed according to the transformed coordinates in $\mathbf{x}^o$ to sample the pixel value for the cropped video $\mathbf{x}^c$.
}
\label{fig:sampler}
\vspace{-4mm}
\end{figure}

With the six parameters, the 3D affine matrix $A_\phi$ is essentially a coordinate transformation function that maps the coordinate system from the cropped video $\mathbf{x}^c$ to the original video $\mathbf{x}^o$ with scaling, rotation, and translation in the spatial dimensions as well as scaling and translation in the temporal dimension. 
Given the homogeneous coordinate $(x_i^c, y_i^c, t_i^c, 1)$ in the cropped videos $\mathbf{x}^c$, its corresponding coordinate $(x_i^o, y_i^o, t_i^o)$ in the original video $\mathbf{x}^o$ can be calculated as follows:
\begin{equation}
    \left(\begin{array}{c}
    x_i^o \\
    y_i^o \\
    t_i^o
    \end{array}\right)=A_{\phi}
    \left(\begin{array}{c}
    x_i^c \\
    y_i^c \\
    t_i^c \\
    1
    \end{array}\right)\ ,
\label{transformation}
\end{equation}
where all the coordinates in both original and cropped videos are normalized, \textit{i.e.}, $\{x_i^o, y_i^o, t_i^o, x_i^c, y_i^c, t_i^c\}\in [-1,1]$.
%
%
The coordinates in $\mathbf{x}^c$ are known, which are uniformly distributed between $[-1, 1]$ according to the resolution of $\mathbf{x}^c$.
For example, to crop a video $\mathbf{x}^c$ with a spatio-temporal resolution of 16$\times$112$^2$, we first generate 16 uniform points between $[-1, 1]$ in the temporal axis and 112 points in two spatial axes. Then the obtained 3D grids with 16$\times$112$^2$ coordinates in $\mathbf{x}^c$ are transformed to $\mathbf{x}^o$ by Equation~\ref{transformation}.
%
Because the transformed coordinates in $\mathbf{x}^o$ may not accurately correspond to the pixel, the pixel values for $(x_i^o, y_i^o, t_i^o)$ are sampled by bilinear interpolation. This cropping process is visualized in Figure~\ref{fig:sampler} for intuitive understanding. Please refer to Appendix for more details.

\noindent\textbf{Generating transformation parameters.} 
To enable this cropping process to be learnable, we employ a multi-layer perceptron to predict the aforementioned six 3D affine transformation parameters by:
\begin{equation}
    \mathbf{v} = \sigma(\mathbf{W}_2 \delta(\mathbf{W}_1 \mathbf{n}))\ ,
    \label{eq:mlp}
\end{equation}
\noindent where $\mathbf{n}\in\mathbb{R}^m$ is a random vector, which can provide diversities for different cropped regions. $\mathbf{W}_1\in\mathbb{R}^{d\times m}, \mathbf{W}_2 \in\mathbb{R}^{6\times d}$ are parameters of the multi-layer perceptron and $\delta$ denotes ReLU activation between two linear layers. The elements in $\mathbf{v}$ corresponds to $[s_p, s_t, \theta, \Delta x, \Delta y, \Delta t]^\top$, which is a vector composed by the six controlling parameters in the 3D affine matrix $A_\phi$. The sigmoid function $\sigma$ is employed to constrain the range of output values to avoid generating meaningless views. 
Moreover, for the cropped region scale $s_p$ and temporal scale $s_t$, a near-zero or zero value indicates that an extremely small cube is cropped, which is meaningless as well and degenerate the learned representation.
Therefore, we set a limited interval for each transformation parameter in $\mathbf{v}$:
\begin{equation}
    \mathbf{v'}
    =\left[\begin{array}{c}
    \check{s_{p}} \\
    \check{s_{t}} \\
    \check{\theta} \\
    \check{\Delta x} \\
    \check{\Delta y} \\
    \check{\Delta t}
    \end{array}\right] + \mathbf{v} \odot \left(
    \left[\begin{array}{c}
    \hat{s_{p}} \\
    \hat{s_{t}} \\
    \hat{\theta} \\
    \hat{\Delta x} \\
    \hat{\Delta y} \\
    \hat{\Delta t}
    \end{array}\right] - 
    \left[\begin{array}{c}
    \check{s_{p}} \\
    \check{s_{t}} \\
    \check{\theta} \\
    \check{\Delta x} \\
    \check{\Delta y} \\
    \check{\Delta t}
    \end{array}\right]
    \right)\ ,
\label{eq:max_min}
\end{equation}
\noindent where $\odot$ indicates the element-wise multiplication and $\check{*}$ and $\hat{*}$ represent the minimum and maximum value allowed during training, respectively. To ensure that the cropped region will always fall within the cube of the original video, the offsets in spatial and temporal need be constrained by: $\check{\Delta x} = \check{\Delta y} = s_p - 1 $, $\hat{\Delta x} = \hat{\Delta y} = 1 - s_p$, $\check{\Delta t} = s_t - 1$ and $\hat{\Delta t} = 1 - s_t$. 
This avoids exceeding the boundary of the original video and yielding invalid views.

Our differentiable 3D affine cropping thus consists of a multi-layer perceptron that generates the transformation parameters and a 3D cubic cropping approach taking these transformation parameters to generate the cropped views. A ParamCrop framework contains two independent affine cropping modules, with one for each view, as in Figure~\ref{fig:fw}.

\subsection{Objective, Optimization and Constraints}
With the 3D affine transformation-based sampling, the cropping operation can be optimized simultaneously with the video backbone in the contrastive training process. Here we introduce the objective, the optimization approach, and some constraints when optimizing for the objective. 

\textbf{Objective: gradual increase of view disparity. }
Recall that the objective of the ParamCrop framework is to control the cropping strategy adaptively.
Inspired by curriculum learning~\cite{bengio2009curriculum_learning}, we aim to crop two views based on the differentiable 3D affine transformation such that the disparity between two cropped views gradually increases along the training process. 
Because the weights of the multi-layer perceptron for generating transformation parameters are randomly initialized, which usually contains small values, it generally maps the random noise vector $\mathbf{n}$ to values around $0$. This means that the initially cropped cubes has a substantial overlap and thus a high similarity.
%
With this initialization condition, we set the goal to be in the opposite direction of the video optimization direction, that is, to be adversarial to the contrastive loss.
\textit{i.e.}, $\vartheta^* = \text{argmax}_\vartheta \mathcal{L}$, where $\vartheta^*$ denotes the optimal solution to the cropping module.
This achieves a gradual increase in the disparity of two augmented views.
The rationale behind this training objective is that two identical views for the same video naturally give the lowest contrastive loss. Therefore, to gradually generate distinct views, ParamCrop needs to gradually increase the contrastive loss. 

\textbf{Optimization: gradient reversal. }To optimize this goal, we apply a simple gradient reversal for the multi-layer perceptron during the back propagation. 
As the backbone aims to learn representations by minimizing the contrastive loss, this reversal operation forces the cropping module to maximize the contrastive loss, so that the disparity between two augmented views can be gradually increased.

\textbf{Constraints: early stopping. }
Blindly maximizing the contrastive loss without any constraint may cause the cropping module to rapidly converge into an extreme position to fulfill the training goal, \textit{e.g., }two diagonal regions in the space-time cube. 
Since it keeps to maximize the contrastive loss, it further yields views with no shared contents until the end of training.
This makes the disparity of two views too large, which is difficult for the model to learn robust representations.
As recent research shows that a sweet spot exists in the intensity of augmentations for a generalized representation~\cite{tian2020whatmakes}, we propose to apply an early stopping strategy to avoid the extreme solutions:
\begin{equation}
    v_i=\left\{
\begin{array}{ccl}
  v_i     &      & {|v_i - 0.5|\leq 0.5-b_{\text{detach}}}\\
\text{detach}(v_i)     &      & {\text{else}}
\end{array} \right \} ,
\label{limit_diversity}
\end{equation}
where $v_i$ is the $i$-th entry of $\mathbf{v}$, and $b_{\text{detach}}\in[0,0.5]$ is the detach bound. 
$b_{\text{detach}}=0$ is equivalent to disabling the early stopping, while $b_{\text{detach}}=0.5$ disables the training for the cropping module. 
With the early stopping strategy, the video backbone can provide the augmented views with more diversities by avoiding saturations.

\section{Experiments}
\noindent\textbf{Training dataset. }We pre-train the models on the training set of Kinetics-400~\cite{carreira2017k400} dataset, containing 240k training videos with each lasting about 10 seconds.

\noindent
\textbf{Pre-training settings.} 
We adopt S3D-G~\cite{xie2018s3dg} and R2D3D~\cite{tran2018r2d3d} as our backbone, and employ SimCLR~\cite{chen2020simclr} and MoCo\cite{he2020moco} as our contrastive learning frameworks. 
The 3D affine cropping module takes $64$ frames with $128^2$ resolution as inputs, and outputs the augmented views with spatial-temporal size $16\times112^2$ to the video backbone.
LARS~\cite{you2017lars} is employed as optimizer.
The batch size, learning rate, and weight decay are set to 1024, 0.3, and 1e-6, respectively. 
Color jittering and random horizontal flip are employed before our cropping module.
Unless otherwise specified, we set the minimum scales $\check{s_p}$ and $\check{s_t}$ to 0.5 and the maximum scales $\hat{s_p}$ and $\hat{s_t}$ to 1.0. 
The detach bound $b_{\text{detach}}$ for early stopping are set to 0.2.
%
%
When compared with other methods, the models are pre-trained with 100 epochs.
For ablation studies, if not specific, we employ S3D-G networks with SimCLR for pre-training, and network are only pre-trained with 20 epochs for efficiency.

\noindent\textbf{Evaluations. }The evaluations of the trained representations are performed on two downstream tasks, \textit{i.e.,} action recognition and video retrieval, on two public datasets: (i) UCF101~\cite{soomro2012ucf101} dataset with 13320 videos from 101 action categories; (ii) HMDB51~\cite{jhuang2011hmdb51} dataset contains 6849 videos from 51 action classes.

\noindent
\textbf{Fully fine-tuning and Linear fine-tuning settings.}
The pre-trained models are fine-tuned on both UCF101 and HMDB51 with a resolution of $32\times224^2$. 
We use Adam~\cite{kingma2014adam} with batch size of $128$ and weight decay 1e-3.
%
The learning rate for fully fine-tuning is set to 0.0002, while 0.002 for linear fine-tuning. 
In the fine-tuning phase, the common data augmentation strategies are adopted, such as color jittering, random cropping and random horizontal flip. 
%

\subsection{Understanding ParamCrop}
\begin{figure}[t]
\begin{center}
   \includegraphics[width=0.95\linewidth]{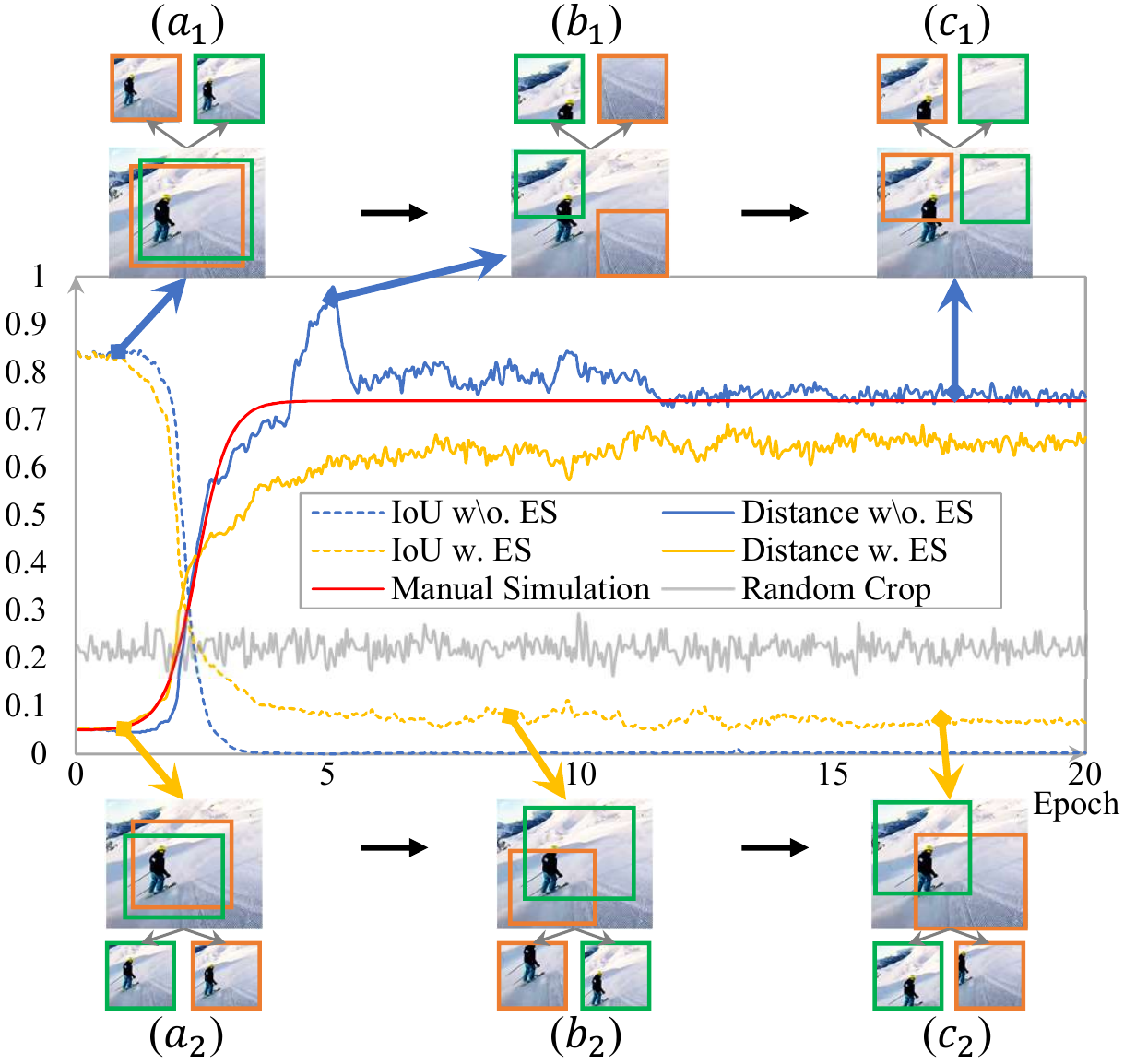}
   \vspace{-4mm}
\end{center}
   \caption{The curve of IoU (dashed) and center Manhattan distance (solid) between two cropping regions generated by ParamCrop along the training process. The yellow and blue curves respectively indicate the training process with/without early stopping.
   The red curve is a smooth center distance curve simulated manually.
    \textbf{($a_1,a_2,b_1,b_2,c_1,c_2$)} show the relative position between the two cropping regions at different stages in training. 
   \textbf{($a_2,b_2,c_2$)} employs early stopping while \textbf{($a_1,b_1,c_1$)} does not.
   Note that although only spatial regions are visualized as an showcase, the similar trend also exists between two views in the temporal dimension. 
   }
   \vspace{-2mm}
\label{fig:iou_dis}
\end{figure}

We first visualize the curve of spatio-temporal IoU and 3D center Manhattan distance between two cropped regions along with the contrastive training in Figure~\ref{fig:iou_dis}. 
Analysis show the following two properties of ParamCrop:

\noindent\textbf{ParamCrop gradually increases the disparity between views.} 
For random cropping, the average distance between two views does not change much with the training process, which indicates the distribution of disparity between views keeps almost unchanged.
For ParamCrop, the disparity gradually increases: 
at the initialization stage, the cropped cubes share a large portion of common visual contents (Figure~\ref{fig:iou_dis} $(a_1)$, $(a_2)$); 
with the gradual increase in the center distance and decrease in the IoU, the amount of shared content gradually decreases, indicating the increase of disparity.

\noindent\textbf{Early stopping ensures reasonable overlap between views. }Without the early stopping strategy, the maximum distance is quickly reached after a short training (diagonal locations for two views in Figure~\ref{fig:iou_dis} $(b_1)$), before the distance reduces to 0.75 and oscillates around it.
This is probably because the learning process without early stopping has a strong momentum, but a distance around 0.75 is sufficiently large to generate two views with no overlap at late stage of training, as in Figure~\ref{fig:iou_dis} $(c_1)$. However, ParamCrop with early stopping can \textit{prevent the extreme locations and box sizes}, which ensures enough shared semantic information between the two views, as in Figure~\ref{fig:iou_dis} $(b_2),(c_2)$.

\subsection{Ablation Studies}
\noindent\textbf{Different cropping strategies.}
To further investigate the improvement brought by ParamCrop, we explore following cropping strategies:

\noindent\textbf{(i) Simple: }Much shared contents as in Figure~\ref{fig:iou_dis}($a_1$);

\noindent\textbf{(ii) Hard: } Less shared contents as in Figure~\ref{fig:iou_dis}($b_1$)\&($c_1$);

\noindent\textbf{(iii) Manual simulation: }
Gradually increasing amount of shared contents, as in Figure~\ref{fig:iou_dis} from $(a_2)$ to $(c_2)$;

\noindent\textbf{(iv) AutoAugment: }
The augmentation searched by~\cite{cubuk2019autoaugment}.

It can be observed in Table~\ref{table:difficulty} that AutoAugment~\cite{cubuk2019autoaugment} achieves similar performance with Random Cropping. Since AutoAugment is designed for supervised training, its essence is a combination of different augmentations, and it dose not consider the training process.
Note that fixing the disparity between two cropping regions too simple or hard, no robust representation can be given.
Meanwhile, if we manually simulate the cropping process as in ParamCrop, we can observe a notable improvement in fully finetuning, but the linear evaluation significantly drops. 
This shows that rigidly generating cropping regions from low disparity to high disparity along the training process can only provide a better initialization for action recognition, but the learnt representation is less general. 
In comparison, ParamCrop yields both better initialization and more generalized representations, since it not only crops views with disparity gradually increasing, but also controls the process adaptively.

\begin{table}[t]
\centering
\small
\tablestyle{2pt}{1.0}
\begin{tabular}{c|ll|ll}
\multirow{2}*{\small Strategies}  & \multicolumn{2}{c|}{\small HMDB51} & \multicolumn{2}{c}{\small UCF101} \\
~  & \multicolumn{1}{c}{Finetune} & \multicolumn{1}{c|}{Linear}  &   \multicolumn{1}{c}{Finetune} & \multicolumn{1}{c}{Linear} \\

\shline
  Random &   56.0 & 33.5  &  85.3  & 57.9 \\
  \tabincell{c}{AutoAugment~\cite{cubuk2019autoaugment}}   &  55.7 {\scriptsize\color{red}{(-0.3)}} & 32.2 {\scriptsize\color{red}{(-1.3)}}    &  86.2 {\scriptsize\color{forestgreen}{(+0.9)}} & 57.2 {\scriptsize\color{red}{(-0.7)}}   \\ 
  \hline
  Simple  &  54.6 {\scriptsize\color{red}{(-1.4)}} & 33.1 {\scriptsize\color{red}{(-0.4)}}    &  85.3 {\scriptsize(+0.0)}  & 53.6 {\scriptsize\color{red}{(-4.3)}}   \\  
  Hard   &   55.2 {\scriptsize\color{red}{(-0.8)}} & 30.1 {\scriptsize\color{red}{(-3.4)}}   &  85.5 {\scriptsize\color{forestgreen}{(+0.2)}}  & 46.7 {\scriptsize\color{red}{(-10.5)}} \\  
  \tabincell{c}{Manual  Simulation}   &  58.6 {\scriptsize\color{forestgreen}{(+2.6)}} & 27.6 {\scriptsize\color{red}{(-5.9)}}    &  86.6 {\scriptsize\color{forestgreen}{(+1.1)}} & 50.3 {\scriptsize\color{red}{(-7.9)}}   \\ 
  \hline
  ParamCrop   &   \textbf{59.9} {\scriptsize\color{forestgreen}\textbf{(+3.9)}}  & \textbf{37.3} {\scriptsize\color{forestgreen}\textbf{(+3.8)}}    &  \textbf{86.9} {\scriptsize\color{forestgreen}\textbf{(+1.6)}}& \textbf{59.3} {\scriptsize\color{forestgreen}\textbf{(+1.4)}}   \\
\end{tabular}
\caption{Comparing different data augmentation strategies.
}
\vspace{-0.8mm}
\label{table:difficulty}
\end{table}

\begin{table}[t]
\centering
\footnotesize
\tablestyle{3pt}{1.0}
\begin{tabular}{c|cc|ll|ll}
\multirow{2}*{\small Rand.} & \multicolumn{2}{c|}{\small P.C.}  & \multicolumn{2}{c|}{\small HMDB51} & \multicolumn{2}{c}{\small UCF101} \\
~ & T. & S. & \multicolumn{1}{c}{Finetune} & \multicolumn{1}{c|}{Linear}  &  \multicolumn{1}{c}{Finetune} & \multicolumn{1}{c}{Linear} \\
\shline
  \cmark  &  \xmark  & \xmark & 56.0 & 33.5  &  85.3 & 57.9 \\
  \xmark &  \cmark  & \xmark & 53.5 {\footnotesize\color{red}{(-2.5)}} & 27.3 {\footnotesize\color{red}{(-6.2)}}  & 84.9 {\footnotesize\color{red}{(-0.4)}}  & 45.0 {\footnotesize\color{red}{(-12.9)}}\\
  \xmark &  \xmark  & \cmark & 59.9 {\footnotesize\color{forestgreen}{(+6.4)}} &  32.7 {\footnotesize\color{red}{(-0.8)}}  & 86.5 {\footnotesize\color{forestgreen}{(+1.2)}} & 41.3 {\footnotesize\color{red}{(-16.6)}}\\
  \xmark &  \cmark  & \cmark & 58.0 {\footnotesize\color{forestgreen}{(+2.0)}} & 37.2 {\footnotesize\color{forestgreen}{(+3.7)}}   &  86.9 {\footnotesize\color{forestgreen}{(+1.6)}} & 58.5 {\footnotesize\color{forestgreen}{(+0.6)}}   \\
  \hline
 \cmark &  \cmark  & \cmark & \textbf{59.9} {\footnotesize\color{forestgreen}\textbf{(+3.9)}}  & \textbf{37.3} {\footnotesize\color{forestgreen}\textbf{(+3.8)}}   & \textbf{86.9} {\footnotesize\color{forestgreen}\textbf{(+1.6)}}  & \textbf{59.3} {\footnotesize\color{forestgreen}\textbf{(+1.4)}}\\
\end{tabular}
\caption{
Decomposing ParamCrop into spatial cropping (Spat.) and temporal cropping (Temp.).
'Rand.' indicates the usage of random cropping in $R_1, R_2$.
For the combination of random cropping and ParamCrop, we insert random cropping before ParamCrop.
}
\label{table:spatial_temporal}
\end{table}

\noindent\textbf{Spatial cropping and temporal cropping.} We decompose ParamCrop into the parametric spatial cropping and the parametric temporal cropping, and evaluate them independently by fixing the other to central crop. The results are shown in Table~\ref{table:spatial_temporal}.
For fully fine-tuning, the parametric temporal cropping yields a slightly lower performance than random cropping, while the parametric spatial cropping performs slightly higher. 
In terms of linear evaluation, both only-spatial and only-temporal ParamCrop underperform the random cropping one.  
However, the full ParamCrop with both spatial and temporal cropping can notably improve both full and linear evaluations.
Additionally, a further improvement is observed by inserting the standard random cropping before ParamCrop, which means that ParamCrop can essentially enhance the existing framework.
For further experiments, unless otherwise stated, both random crop and ParamCrop are used.

\noindent\textbf{Spatial aspect ratio and rotation.} Adjusting spatial aspect ratio and rotation angle can be realized by affine transformation, which are controlled by the region scale $s_p$ and rotation angle $\theta$ in transformation matrix. However, as shown in Table~\ref{table:rot_aspect}, introducing aspect ratio and rotation to ParamCrop has an adverse effect on the representation. Since there is less abnormal scaling and rotation in natural videos. This is in line with previous research findings~\cite{chen2020simclr}. Hence, we disable aspect ratio learning and set the rotation angle $\check{\theta}$ and $\hat{\theta}$ both to 0.0 in other experiments.

\begin{table}[t]
\centering
\footnotesize
\tablestyle{4pt}{1.0}
\begin{tabular}{c|c|ll|ll}
 \multirow{2}*{\tabincell{c}{\small Method}} &  \multirow{2}*{\tabincell{c}{\small G.R.}} &  \multicolumn{2}{c|}{\small  HMDB51} & \multicolumn{2}{c}{\small  UCF101}  \\
 ~ & ~  &  \multicolumn{1}{c}{\small Finetune}  & \multicolumn{1}{c|}{Linear}  &  \multicolumn{1}{c}{Finetune} & \multicolumn{1}{c}{Linear} \\
\shline
\small R.C. & - & 56.0 & 33.5 & 85.3 & 57.9 \\
\hline
 \multirow{2}*{\small P.C.}  & \xmark   &   56.0  & 33.2   &     84.3 & 55.2   \\ 
~  &\cmark   &   \textbf{59.9} {\scriptsize\color{forestgreen}\textbf{(+3.9)}} & \textbf{37.3} {\scriptsize\color{forestgreen}\textbf{(+3.8)}}     &  \textbf{86.9} {\scriptsize\color{forestgreen}\textbf{(+1.6)}} & \textbf{59.3} {\scriptsize\color{forestgreen}\textbf{(+1.4)}} \\
\end{tabular}
\caption{The importance of the gradient reversal (G.R.) operation for ParamCrop (P.C.). R.C. refers to random cropping.
}
\vspace{-1mm}
\label{table:gradient_reverse}
\end{table}

\begin{table}[t]
\centering
\footnotesize
\tablestyle{6pt}{1.0}
\begin{tabular}{c|cc|cc}
\multirow{2}*{\tabincell{c}{\small Early  Stopping}} &  \multicolumn{2}{c|}{\small  HMDB51} & \multicolumn{2}{c}{\small  UCF101}  \\
~ & Finetune  & Linear  &  Finetune & Linear \\
\shline
\xmark   &   59.9  &  32.6 &  \textbf{87.0} &   55.8     \\
\cmark   &   59.9   &  \textbf{37.3} & 86.9  &  \textbf{59.3}  \\
\end{tabular}
\caption{Ablation on the early stopping strategy. 
}
\label{table:diversity_limitation}
\vspace{-1mm}
\end{table}

\begin{table}[t]
\centering
\footnotesize
\tablestyle{6pt}{1.0}
\begin{tabular}{cc|cc|cc}
\multirow{2}*{\tabincell{c}{\small A.R.}} & \multirow{2}*{\tabincell{c}{\small Rot.}} &  \multicolumn{2}{c|}{\small  HMDB51} & \multicolumn{2}{c}{\small  UCF101}  \\
~ & ~ & Finetune  & Linear  &  Finetune & Linear \\
\shline
\cmark   &  \xmark   &   55.7  &  34.7 &  84.5 &   53.6     \\
\xmark   &  \cmark   &   55.8  &  29.5 &  86.2 &   48.6     \\
\xmark   &  \xmark   &   \textbf{59.9}  &  \textbf{37.3} & \textbf{86.9}  &  \textbf{59.3}  \\
\end{tabular}
\caption{Exploring two spatial cropping items, \ie, Aspect Ratio (A.R.) and Rotation (Rot.) in ParamCrop.
}
\label{table:rot_aspect}
\vspace{-1mm}
\end{table}

\begin{figure}[t]
\begin{center}
  \includegraphics[width=1\linewidth]{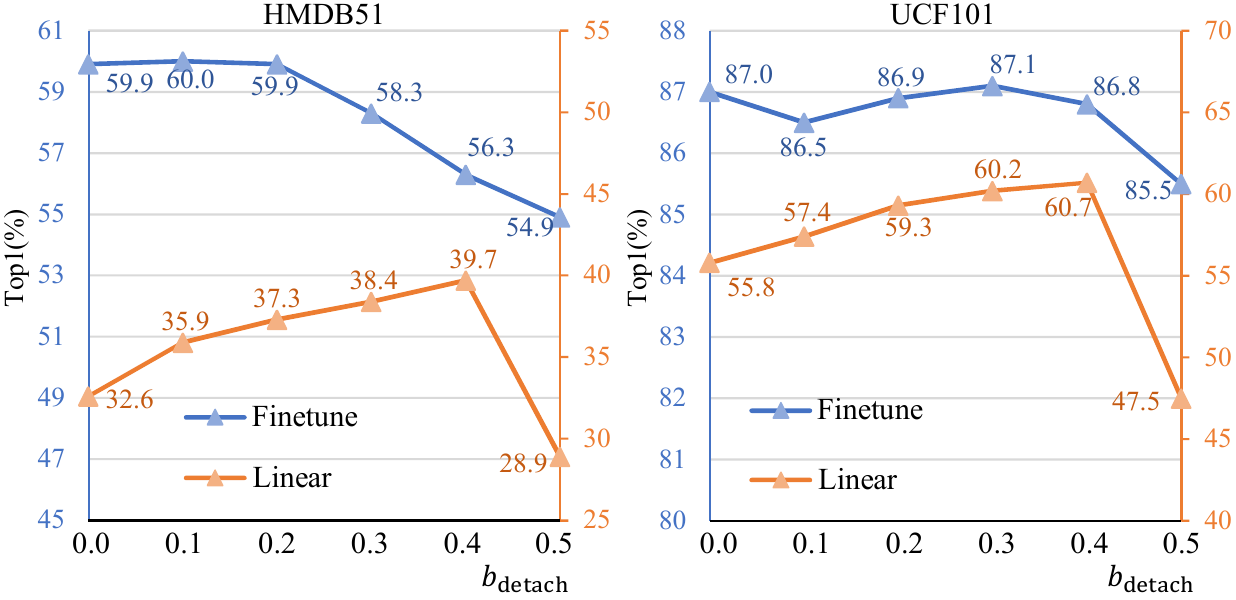}
\end{center}
  \vspace{-3mm}
  \caption{Sensitivity analysis for the detach bound $b_\text{detach}$ in early stopping on both HMDB51 and UCF101.}
  \vspace{-3mm}
\label{fig:v_l_v_u}
\end{figure}

\noindent\textbf{Gradient reversal.} 
Table~\ref{table:gradient_reverse} shows the importance of gradient reversal for ParamCrop.
Removing gradient reversal makes the objective of ParamCrop identical to the backbone, \textit{i.e.,} to minimize the contrastive loss. 
Hence, in this case, ParamCrop trys to increase the amount of shared visual contents between two spatio-temporal cubes cropped by the cropping module, which encourages the model to find shortcuts, yielding sub-optimal representations.

\noindent \textbf{Early stopping.} Qualitatively, it is observed in Figure~\ref{fig:iou_dis} ($a_2$-$b_2$-$c_2$) that early stopping avoids extreme locations (compared to Figure~\ref{fig:iou_dis} ($a_1$-$b_1$-$c_1$) with no early stopping), which ensures shared contents for cropped regions. Quantitatively, the results in Table~\ref{table:diversity_limitation} shows that early stopping notably increases the linear separability of the learned representation, while it has less effect on the fully fine-tuning.

\noindent\textbf{Sensitivity analysis of the detach bound $b_\text{detach}$.}
As in Figure~\ref{fig:v_l_v_u}, because of the decreased disparity, the performance of fully fine-tuning gradually drops with the increase of the detach bound. 
However, since there exists a sweet spot of the intensity of the augmentations in contrastive learning for the representation quality~\cite{tian2020whatmakes}, the performance of linear evaluation gradually increases before it drops.
To strike the balance between linear and full evaluations, we choose $0.2$ as our detach bound.

\noindent\textbf{Different frameworks and backbones.} 
To evaluate the applicability of the proposed approach, ParamCrop is integrated to two mainstream contrastive learning frameworks SimCLR~\cite{chen2020simclr} and MoCo~\cite{he2020moco} and two common video backbones S3D-G~\cite{xie2018s3dg} and R-2D3D~\cite{tran2018r2d3d}.
The results in Table~\ref{table:general} demonstrate that ParamCrop can be generalized to multiple contrastive frameworks and video backbones. 

\noindent\textbf{Data efficiency.} Table~\ref{table:data_size} shows that as we reduce the number of labelled training data during fine-tuning, ParamCrop consistently outperforms random cropping, which shows that the representation yielded by ParamCrop has a higher data efficiency. 

\noindent\textbf{Semi-supervised learning.}
As a popular method in the semi-supervised task, Mean Teacher~\cite{tarvainen2017mean_teacher} also contains a contrastive learning branch similar to MoCo~\cite{he2020moco}. 
Table~\ref{table:mean_teacher} shows ParamCrop can also replace the random cropping in the semi-supervised settings to enhance the performance. 

\begin{table}[t]
\centering
\footnotesize
\tablestyle{4.5pt}{1.0}
\begin{tabular}{cc|c|ll}
\small Framework & \small Architecture  & \small P.C.     & \multicolumn{1}{c}{HMDB51}   &  \multicolumn{1}{c}{UCF101} \\
\shline
SimCLR    & S3D-G     &  \xmark   &  56.0  &  85.3 \\ 
SimCLR    & S3D-G     &  \cmark   &  \textbf{59.9} {\color{forestgreen}\scriptsize\textbf{(+3.9)}} &  \textbf{86.9} {\color{forestgreen}\scriptsize\textbf{(+1.6)}}\\
SimCLR    & R-2D3D     &  \xmark   &  50.4  &  77.2 \\  
SimCLR    & R-2D3D     &  \cmark   &  \textbf{53.0} {\color{forestgreen}\scriptsize\textbf{(+2.6)}}  &  \textbf{79.4} {\color{forestgreen}\scriptsize\textbf{(+2.2)}} \\  
\hline
MoCo    & S3D-G     &  \xmark   &  52.4  &  84.1 \\
MoCo    & S3D-G     &  \cmark   &  \textbf{54.3} {\color{forestgreen}\scriptsize\textbf{(+1.9)}}  &  \textbf{85.1} {\color{forestgreen}\scriptsize\textbf{(+1.0)}} \\
MoCo    & R-2D3D     &  \xmark   &  45.4  &  72.8 \\
MoCo    & R-2D3D     &  \cmark   &  \textbf{48.2} {\color{forestgreen}\scriptsize\textbf{(+2.8)}} &  \textbf{73.8} {\color{forestgreen}\scriptsize\textbf{(+1.0)}} \\

\end{tabular}
\caption{Integrating ParamCrop (P.C.) with different contrastive frameworks and video backbones. The fully fine-tuning performances are reported here.
}
\label{table:general}
\vspace{-1mm}
\end{table}

\begin{table}[t]
\begin{center}
\footnotesize
\tablestyle{7pt}{1.0}
\begin{tabular}{c|c|cc|cc}
\multirow{2}*{\small Architecture}  & \multirow{2}*{\small Label} &  \multicolumn{2}{c|}{\small HMDB51}   &  \multicolumn{2}{c}{\small UCF101} \\
~  & ~ &  R.C.   &  P.C. &  R.C.   &  P.C.\\
\shline
 \multirow{3}*{S3D-G}  &  50\%  &  51.5  &  \textbf{52.8} & 82.4 & \textbf{83.3}\\
~  &  30\% &  44.9  &  \textbf{46.0} & 76.9 & \textbf{78.5} \\
~  &  10\%   &   33.3  &  \textbf{34.6}& 57.6 & \textbf{60.1} \\
\end{tabular}
\end{center}
\vspace{-2mm}
\caption{Evaluating random cropping (R.C.) and ParamCrop (P.C.) by fully fine-tuning pre-trained models with less labelled data on HMDB51 and UCF101.
}
\label{table:data_size}
\end{table}

\begin{table}[t]
\centering
\footnotesize
\tablestyle{6pt}{1.0}
\begin{tabular}{c|c|cc}
\small Architecture  & \small CroppingMethod &  \small 10\% Label   &  \small 50\% Label \\
\shline
 \multirow{2}*{\small S3D-G}  &  RandomCrop  &  13.9  &  29.9 \\
 ~  &  ParamCrop &  \textbf{14.4}  &  \textbf{31.6} \\
\end{tabular}
\caption{
Applying ParamCrop to the semi-supervised task.
All experiments are based on the Mean Teacher~\cite{tarvainen2017mean_teacher} framework on HMDB51 dataset, and the models are initialized randomly. }
\label{table:mean_teacher}
\vspace{-1mm}
\end{table}

\subsection{Comparison with the Existing Approaches}
\begin{table}[t]
\centering
\footnotesize
\tablestyle{2.5pt}{1.05}
\begin{tabular}{c|ccc|cc}
\small Approach & Architecture & \tabincell{c}{Pre-train \\ Dataset} & Res. & HMDB & UCF   \\
\shline
\rowcolor{grey} \greyColor{$\rho$SimCLR}~\cite{feichtenhofer2021largescale} & \greyColor{R3D-50} & \greyColor{K400} & \greyColor{224} & \greyColor{-} & \greyColor{88.9} \\
\rowcolor{grey} \greyColor{CVRL}~\cite{qian2021cvrl} & \greyColor{R3D-50} & \greyColor{K400} & \greyColor{224}  & \greyColor{66.7} & \greyColor{92.2} \\
VCOP~\cite{xu2019vcop} &  R(2+1)D-10 & UCF                    & 112 & 30.9  & 72.4 \\
DPC~\cite{han2019dpc}    &    R-2D3D-18   &        K400       & 224     &    35.7  &    75.7\\
CBT~\cite{sun2019cbt}       &  S3D-23    & K600               & 112 &  44.6   &  79.5 \\
MemDPC~\cite{han2020memdpc} &    R-2D3D-18          &   K400  & 224 &  41.2    & 78.1   \\
SpeedNet~\cite{benaim2020speednet}    &    S3D-G-23  &  K400  & 224 &   48.8   &  81.1  \\
DynamoNet~\cite{diba2019dynamonet}    &    STCNet  &Y8M    & 224 &   59.9   &  88.1  \\
DSM~\cite{wang2020dsm} & R3D-34 & K400                   & 224 &  52.8 & 78.2 \\
MoSI~\cite{mosi} & R2D3D-18 & K400                            & 112 &     48.6 & 70.7 \\
MLRep~\cite{qian2021mlrep} & R3D-18 & K400                   & 112 &     47.6 & 79.1 \\
RSPNet*~\cite{chen2020rspnet} & S3D-G-23 & K400               & 224 & 59.6 & 89.9 \\
STS*~\cite{wang2021sts} & S3D-G-23 & K400                     & 224 &  62.0 & 89.0 \\
\hline
ParamCrop   &    R-2D3D-18    &        K400     & 224 &    53.7  &   82.8 \\
ParamCrop   &    S3D-G-23     &        K400     & 112 &  51.4    &  80.2   \\
ParamCrop   &    S3D-G-23     &        K400     & 224 &  62.3    &  88.9   \\
ParamCrop*   &    S3D-G-23     &        K400     & 224 &  \textbf{63.4}    &  \textbf{91.3}   \\
\hline 
\rowcolor{grey} \greyColor{Supervised}~\cite{xie2018s3dg}   &   \greyColor{S3D-G-23}     &        \greyColor{K400}     &  \greyColor{224} &  \greyColor{75.9}    &  \greyColor{96.8}   \\
\end{tabular}
\caption{Comparison with existing methods, where pre-trained models are fully fine-tuned in HMDB51 and UCF101. For pre-train dataset, `K400' and `K600' refer to Kinetics-400 and Kinetics-600~\cite{carreira2018k600} datasets, `Y8M' is the Youtube8M dataset. `*' indicates that 64 frames are used to fine-tune backbone. `Res.' is the spatio resolution used in fine-tuning.}
\label{table:sota_ft}
\end{table}

\begin{table}[t]
\begin{center}
\tablestyle{2.5pt}{1.05}
\begin{tabular}{c|cc|cc}
 Approach & Architecture & \tabincell{c}{Pre-train \\ Dataset}  &   HMDB51 & UCF101   \\
\shline
MemDPC~\cite{han2020memdpc} &    R-2D3D-18    &        K400     &  30.5 & 54.1   \\
MemDPC n.l.~\cite{han2020memdpc} &    R-2D3D-18    &        K400     & 33.6 & 58.5    \\
CBT~\cite{sun2019cbt} &    S3D-23    &        K600     & 29.5 & 54.0    \\
MLRep~\cite{qian2021mlrep} &    R3D-18    &        K400     & 33.4 & 63.2    \\
\hline
ParamCrop   &    R-2D3D-18    &        K400     &   \textbf{39.7}   &  66.8  \\
ParamCrop   &    S3D-G-23     &        K400     &  39.4    &  \textbf{68.5}   \\
\end{tabular}
\end{center}
\vspace*{-3mm}
\caption{Comparison with existing methods in linear evaluation on HMDB51 and UCF101. `n.l.' refers to a nonlinear classifier.}
\label{table:sota_lft}
\vspace*{-1mm}
\end{table}

\noindent\textbf{Action Recognition.} 
Table~\ref{table:sota_ft} compares ParamCrop existing methods under full fine-tune setting on HMDB51 and UCF101.
In these experiments, we choose SimCLR~\cite{chen2020simclr} as the base contrastive learning framework.
From these results, we can draw following conclusions:
\textit{(i)} ParamCrop achieves remarkable performance on the two datasets. 
Especially, we surpass SpeedNet~\cite{benaim2020speednet} by $13.5\%$ and $7.8\%$ on the two datasets respectively, when applying same dataset and backbone, \emph{i.e.}, Kinetics 400 and S3D-G;
Even with small resolution in fine-tuning, we also surpass MLRep~\cite{qian2021mlrep} by 4.8\% on HMDB.
\textit{(ii)} 
ParamCrop can be trained on less data but achieve remarkable performance. 
DynamoNet~\cite{diba2019dynamonet} is trained with 8M videos on Youtube8M, while ParamCrop just uses 240K videos but still obtains $2.4\%$ and $0.8\%$ gains;
\textit{(iii)} 
ParmaCrop is lightly lower than CVRL~\cite{qian2021cvrl}, which may because it applies a deeper backbone (\ie, R3D-50) and larger pre-training resolution (\ie $16\times224^2$, but we still obtain competitive performance on UCF101.
Table~\ref{table:sota_lft} compares the linear evaluation on HMDB51 and UCF101.
Compared with MemDPC~\cite{han2020memdpc}, ParamCrop respectively gains $6.1\%$ and $8.3\%$ on HMDB51 and UCF101 using R-2D3D, demonstrating ParamCrop can learn powerful video representations.

\noindent\textbf{Video Retrieval.} 
In the video retrieval task, we mainly follow the settings in previous works~\cite{xu2019vcop,luo2020vcp,han2020memdpc}.
The models pre-trained by ParamCrop on Kinetics400 using SimCLR are employed as the feature extractor without fine-tuning.
We conduct experiments on both HMDB51 and UCF101, and compare with other approaches in Table~\ref{table:retrieval_hmdb} and Table~\ref{table:retrieval_ucf} respectively. 
Results show that ParamCrop exceeds the MemDPC~\cite{han2020memdpc} by $6.4\%$ and $2.8\%$ with the same backbone on HMDB51 and UCF101, respectively, which indicates that our learned representation is more generalized.

\begin{table}[t]
\centering
\footnotesize
\tablestyle{4pt}{1.0}
\begin{tabular}{cc|cccc}
\small Approach  & Architecture &  R@1 & R@5 &  R@10  &  R@20 \\
\shline
VCOP~\cite{xu2019vcop} & R3D-18 & 7.6 & 22.9  & 34.4 & 48.8 \\
VCP~\cite{luo2020vcp} & R3D-18 & 7.6 & 24.4 & 36.3 & 53.6 \\
 DSM~\cite{wang2020dsm} & C3D-34 & 8.2 &  25.9 & 38.1 & 52.0 \\
 MemDPC~\cite{han2020memdpc} & R-2D3D-18 & 15.6 &  37.6 & 52.0 & 65.3 \\
 \shline
ParamCrop & R-2D3D-18  & \textbf{21.9} & \textbf{46.9} & 59.0 & 71.5 \\
ParamCrop  &  S3D-G-23 & 23.3 & 46.8 & 59.4 & 72.8 \\

\end{tabular}
\caption{Nearest neighbour retrieval comparison on HMDB51.}
\label{table:retrieval_hmdb}
\vspace{-1.5mm}
\end{table}

\begin{table}[t]
\centering
\footnotesize
\tablestyle{4pt}{1.0}
\begin{tabular}{cc|cccc}
\small Approach  & Architecture & R@1 & R@5 &  R@10  &  R@20 \\
\shline
Jigsaw~\cite{noroozi2016jigsaw} & \small CFN & 19.7 & 28.5 & 33.5 & 40.0 \\
OPN~\cite{lee2017sort_seq} & \small OPN & 19.9 & 28.7 & 34.0 & 40.6 \\
VCOP~\cite{xu2019vcop} & \small R3D-18 & 14.1 & 30.3 & 40.4 & 51.1 \\
VCP~\cite{luo2020vcp} & \small R3D-18 & 18.6 & 33.6  & 42.5 & 53.5 \\
 SpeedNet~\cite{benaim2020speednet} & \small S3D-G-23 & 13.0 &  28.1 & 37.5 & 49.5 \\
 DSM~\cite{wang2020dsm} &\small C3D-34 & 16.8 &  33.4 & 43.4 & 54.6 \\
 MemDPC~\cite{han2020memdpc} &\small R-2D3D-18 & 40.2 &  63.2 & 71.9 & 78.6 \\
 \shline
ParamCrop &\small R-2D3D-18  & 43.0 & 59.9 & 69.2 & 78.3 \\
ParamCrop  &\small  S3D-G-23 &  \textbf{46.3}   &  \textbf{62.3}  & \textbf{71.3}  & \textbf{79.1} 
\end{tabular}
\caption{Nearest neighbour retrieval comparison on UCF101.}
\label{table:retrieval_ucf}
\vspace{-5mm}
\end{table}

\section{Conclusion}
In this work, 
we propose a parametric cubic cropping for adaptively controlling the disparity between two cropped views along the training process. 
Specifically, we enable online training by first extending the affine transformation matrix to the 3D affine transformation and learn to regress the transformation parameters such that the cropping operation is fully differentiable. 
%
%
For the optimization, the parametric cubic cropping operation is trained with an adversarial objective to the video backbone, and optimized using a simple gradient reversal operation. We additionally show that an early stopping strategy in the optimization process is beneficial for the video backbone to learn more robust representations. 
Empirical results demonstrate that the adaptively controlled disparity between views is indeed effective for improving the representation quality. 
Extensive ablation studies validate the effectiveness of each proposed component and evaluations are performed on both action recognition and video retrieval.

\noindent\textbf{Limitations.} 
To achieve flexible spatio-temporal cropping, ParamCrop needs to crop cubes from larger inputs. 
Since the operation is performed on GPU, utilizing ParamCrop would to some extent increase the memory usage of GPUs. 

{\small
\bibliographystyle{ieee_fullname}
\bibliography{egbib}
}

\end{document}